\documentclass[authoryear]{eshbaugh}

\usepackage[utf8]{inputenc}
\usepackage[T1]{fontenc}
\usepackage{hyperref}
\usepackage{url}
\usepackage{amsfonts}
\usepackage{graphicx}
\usepackage{nicefrac}
\usepackage{microtype}
\usepackage{listings}
\usepackage{xcolor}
\usepackage{colortbl}
\usepackage{adjustbox}
\usepackage{array}

\usepackage[acronym, nonumberlist]{glossaries-extra}
\makeglossaries
\glsdisablehyper
\setabbreviationstyle[acronym]{long-short}

\usepackage{tabularx,booktabs,siunitx}
\usepackage{pgffor}
\hypersetup{hidelinks}

\lstdefinelanguage{json}{
  basicstyle=\ttfamily\footnotesize,
  numbers=left,
  numberstyle=\tiny\color{gray},
  stepnumber=1,
  numbersep=5pt,
  showstringspaces=false,
  breaklines=true,
  frame=single,
  backgroundcolor=\color{gray!10},
  string=[s]{"}{"},
  morestring=[b]',
  literate=
   *{0}{{{\color{blue}0}}}{1}
    {1}{{{\color{blue}1}}}{1}
    {2}{{{\color{blue}2}}}{1}
    {3}{{{\color{blue}3}}}{1}
    {4}{{{\color{blue}4}}}{1}
    {5}{{{\color{blue}5}}}{1}
    {6}{{{\color{blue}6}}}{1}
    {7}{{{\color{blue}7}}}{1}
    {8}{{{\color{blue}8}}}{1}
    {9}{{{\color{blue}9}}}{1}
    {:}{{{\color{black}:}}}{1}
    {,}{{{\color{black},}}}{1}
    {"}{{{\color{red}"}}}{1},
}
\usepackage{float}
\usepackage{xcolor}  
\usepackage{amsmath}
\usepackage{subcaption}
\usepackage{placeins}
\usepackage{graphicx}
\usepackage{listings}
\usepackage{pdflscape}

\usepackage{amssymb}
\usepackage{amsmath}

\newglossaryentry{gpt}{
    name = GPT,
    description = {A popular large language model by OpenAI with a wide variety of uses.}
}

\newglossaryentry{llava}{
    name = LLaVA,
    description = {An LLM integrated with CLIP, OpenAI's visual model. Introduced by Liu \textit{et al.} in 2023.}
}

\newglossaryentry{energyplus}{
    name = EnergyPlus,
    description = {A versatile building energy modeling tool provided by the United States Department of Energy. Originally published in 2002.}
}

\newglossaryentry{occlusion}{
    name = Occlusion,
    description = {A test commonly used to evaluate visual machine learning models by blocking out parts of an image and comparing to a baseline.}
}

\newglossaryentry{ablation}{
    name = Ablation,
    description = {A test used to evaluate multimodal models by observing how a model's output responds to each varying modality.}
}

\newacronym{ai}{AI}{Artificial Intelligence}
\newacronym{ml}{ML}{Machine Learning}
\newacronym[plural=LLMs, firstplural={Large Language Models}]{llm}{LLM}{Large Language Model}
\newacronym{idf}{IDF}{Input Definition File}
\newacronym{cop}{COP}{Coefficient of Performance}
\newacronym{json}{JSON}{JavaScript Object Notation}
\newacronym{jsonl}{JSONL}{JSON Lines}
\newacronym{geojson}{GeoJSON}{Geographic JSON}

\newif\ifmarked
\markedfalse

\ifmarked
    \newcommand{\rev}[1]{\textcolor{blue}{#1}}

    \newenvironment{revision}
        {\begingroup\color{blue}\arrayrulecolor{blue}}
        {\endgroup\arrayrulecolor{black}}

    \newcommand{\revtable}[1]{%
        \begingroup
        \color{blue}
        \arrayrulecolor{blue}
        #1
        \arrayrulecolor{black}
        \endgroup
    }
\else
    \newcommand{\rev}[1]{#1}

    \newenvironment{revision}
        {}
        {}

    \newcommand{\revtable}[1]{#1}
\fi

\title{Synthetic Homes: A Multimodal Generative AI Pipeline for Residential Building Data Generation under Data Scarcity}

\shorttitle{Synthetic Homes}

\author{Jackson Eshbaugh}
\affiliation{
Department of Computer Science\\
Lafayette College\\
Easton, PA, USA
}
\email{eshbaugj@lafayette.edu}
\orcid{0009-0009-1806-2166} 

\author{Chetan Tiwari}
\affiliation{
Department of Computer Science\\
Georgia State University\\
Atlanta, GA, USA
}
\email{ctiwari@gsu.edu}
\orcid{0000-0002-3869-4023}

\author{Jorge Silveyra}
\affiliation{
Department of Computer Science\\
Lafayette College\\
Easton, PA, USA
}
\email{silveyrj@lafayette.edu}
\orcid{0009-0004-9769-7250}

\publicationstatus{Preprint}
\version{6.0}
\preprintdate{July 2026}


\begin{document}
\maketitle

\begin{abstract}
Computational models have emerged as powerful tools for multi-scale energy modeling research at the building and urban scale, supporting data-driven analysis across building and urban energy systems. However, these models require large amounts of building parameter data that is often inaccessible, expensive to collect, or subject to privacy constraints. \rev{We introduce a modular framework that applies generative \gls{ai} to construct simulation-ready building datasets from publicly available records and imagery. To improve the reliability of this framework, we evaluate both its AI components and its overall result.} 
Our \rev{occlusion} analysis demonstrates \rev{that for our selected images, \gls{llava}} achieves greater visual focus than a \gls{gpt}-based alternative for building image processing. We also assess \rev{plausibility} of our results against a national reference dataset, finding that our synthetic data overlaps more than 95\% for three of the four selected variables. This work \rev{aims to reduce} dependence on costly or restricted data sources, lowering barriers to building-scale energy research and \gls{ml}-driven urban energy modeling, \rev{thereby providing simulation-ready datasets intended to support downstream applications such as energy modeling, retrofit analysis, and urban-scale simulation under data scarcity.}
\end{abstract}

\section{Introduction}

Consumption of electricity in the United States will increase by \( 1.7\% \) between 2020 and 2026~\parencite{EIA_ElectricityGrowth2025} and seasonal electricity use per person has nearly doubled from 1973 to 2025~\parencite{u.s.environmentalprotectionagencyClimateChangeIndicators2025}. 
Rising energy consumption exerts increasing pressure on infrastructure capacity, operational costs, and greenhouse gas emissions, resulting in rising costs for governments, energy producers, and energy consumers. 
Accurate modeling and policy tools are therefore essential for effective planning, helping prevent infrastructure strain, optimize energy costs, minimize or mitigate climate impacts, assess policy effectiveness, and guide decision-making in urban centers~\parencite{bompardNewElectricityInfrastructure2024, pereraClimateResilientInterconnected2021, zengReviewOptimizationModeling2011, hajriDatadrivenModelHeat2025}.

Computational approaches to monitor energy usage and model urban centers, buildings, and other structures require large amounts of data, such as type and quality of materials, floor plans, and weather patterns to produce scalable and quantitative results~\parencite{u.s.departmentofenergyGettingStarted2025}. However, several issues arise when attempting to collect such information, including high prices, limited availability, and privacy concerns. 
These issues create barriers to accessing data, hindering the development of accurate models and the scaling of models to larger populations or geographic areas. 
The inaccessibility of this data has led researchers to generate synthetic data through a variety of methods.


\noindent The primary contributions of this work are:
\begin{itemize}
    \item \rev{A modular framework that applies multimodal generative \gls{ai} to residential building energy modeling by integrating publicly available records, imagery, and physics-based simulation into a unified workflow for producing simulation-ready residential datasets};
    
    \item \rev{An end-to-end implementation demonstrating that the framework generates 247 synthetic simulation-ready homes at a worst-case cost of \$0.0039 per home while reducing the expertise and data-access barriers associated with residential energy modeling};
    
    \item An occlusion-based analysis approach that provides insight into visual feature usage in language models for building understanding; and
    
    \item A quantitative evaluation demonstrating that the generated data \rev{is plausible} when \rev{evaluated against} ResStock~\parencite{ResStock2024_1Dataset}.
\end{itemize}


 In the following sections, we position our work in the existing literature (Section~\ref{sec:rel_works}) and provide an in depth view of our framework (Section~\ref{sec:pipeline}). Then, we evaluate the \rev{plausibility} of data generated by our pipeline (Section~\ref{sec:validating_data}) \rev{and} we conclude and highlight future directions from this work (Section~\ref{sec:conclusion}).

\section{Related Work}
\label{sec:rel_works}

Current data generation efforts span a wide range of approaches, from optimizing building construction processes to the creation of digital twins. 
For example, \textcite{wanDeepLearningApproach2022} built a generative model that improves the efficiency of energy consumption by determining the optimal layout of floor plans. 
\textcite{elnabawiMethodologyCreatingSynthetic2022} enhanced their building energy models using highly detailed microclimate models that considered features such as shadow generation and seasons to generate weather data. 
\textcite{leeNeuralNetworkBasedBuildingEnergy2019} train a neural network that predicts building energy consumption, aiming to create a more robust model with limited real-world measurements. The network was trained with data synthesized using, among others, electricity, temperature, and humidity, encoded into receiver operating characteristic (ROC) plots which visually encode the separation between classes and are commonly used to assess model discrimination~\parencite{zweigReceiveroperatingCharacteristicROC1993, janssensReflectionModernMethods2020}. 

Other researchers focus on creating and applying digital twins: virtual, dynamic, and data-driven digital replicas of real-world urban environments. 
\textcite{stinnerAutomaticDigitalTwin2021} generate digital twins of building energy systems from piping and instrumentation diagrams using convolutional networks and other algorithms. \textcite{agostinelliCyberPhysicalSystemsImproving2021} create digital twins of energy efficient buildings in residential areas from Building Information Modeling (BIM), Internet of Things (IoT), Geographical Information Systems (GIS), and \gls{ai} components, with machine learning serving as the central module of the model. \textcite{franciscoSmartCityDigital2020} take a different approach by generating digital twins of daily, time-segmented energy benchmarks solely from smart meter data streams. \textcite{belikImplementationDigitalTwin2023} present a hybrid digital twin framework for renewable energy sources (RES) that integrates real-time data, 3D modeling, and sensitivity analysis to improve forecasting and operational efficiency, particularly in grid systems with high RES penetration.

While prior work has focused on traditional simulation tools or sensor data to construct digital twins, \textcite{xuLeveragingGenerativeAI2024} encourage the use of generative \gls{ai} models, such members of the \gls{gpt} family, because they present new opportunities in urban data generation.  For instance, \textcite{dodgeParsingFloorPlan2017} employ convolutional neural networks (CNNs) to parse floor plan images, generating structured representations using wall segmentation, object recognition, and optical character recognition. Relatedly, \textcite{zhangAutomaticBuildingEnergy2025} implement a pipeline that generates \gls{energyplus} IDF files from initial building geometry and textual building descriptions. Moreover, \textcite{xiaoExploringAutomatedEnergy2024} propose a multi-agent \gls{llm} framework that processes unstructured building data and sensor logs to support automated energy optimization in smart grid contexts. Additionally, in closely related fields, generative \gls{ai} has been successfully employed to perform a variety of tasks, such as predictive analysis of traffic congestion~\parencite{linApplicationGenerativeAI2025}, real-time decision-making in urban air mobility~\parencite{shaGenerativeAIEnabledSensing2024}, and stochastic solar irradiance forecasting for building facades~\parencite{zhangSolarGANSyntheticAnnual2022}.

\textcite{liuLargeLanguageModels2025} present a wide-ranging review on the integration of generative \gls{ai} into building energy workflows, highlighting both opportunities and challenges. These include concerns such as dataset preparation, hallucinations, and a lack of domain-specific expertise. However, the impact of such challenges varies depending on the pipeline design. Because our framework combines pretrained \glspl{llm} with simulation tools for quantitative evaluation, we avoid the need for fine-tuning or physics-aware generation, thereby reducing the impact of these concerns. \rev{Moreover,} we evaluate our framework's \gls{llm} components to \rev{avoid generating data impacted by} hallucinations. Additionally, Liu emphasizes the importance of multimodal integration---where multiple data modalities (such as image, text, and tabular data) are used by a process or model. Producing high-quality structured urban data suitable for tasks such as modeling and simulation requires combining multiple complementary processes into a single workflow~\parencite{rehmannEnhancingUrbanEnergy2025, guoCombinedWorkflowGenerate2023, bishopReviewMultiDomainUrban2024}. We adopt these principles in the design of our general framework.


\rev{While prior work has explored using individual and joint components of synthetic building data generation, our contribution is a modular framework that composes a unified workflow for generating synthetic simulation-ready residential energy datasets from publicly available information.} By combining assessor records, street-level imagery, multimodal foundation models (\gls{llava} and \gls{gpt}), and EnergyPlus-compatible output generation, the framework \rev{could substantially reduce} the expertise, data acquisition, and financial barriers traditionally associated with residential energy modeling.

The framework also produces multimodal outputs, including both structured building characteristics suitable for simulation and natural-language inspection \rev{notes} that can \rev{be used for} future machine learning and decision-support tasks. Collectively, these capabilities make it possible to generate complex residential energy datasets through a largely automated workflow, enabling broader participation in building energy research in settings where detailed building inventories are unavailable or prohibitively expensive to obtain.

\section{Methodology}
\label{sec:pipeline}

\begin{figure}
    \centering
    \includegraphics[width=0.75\linewidth]{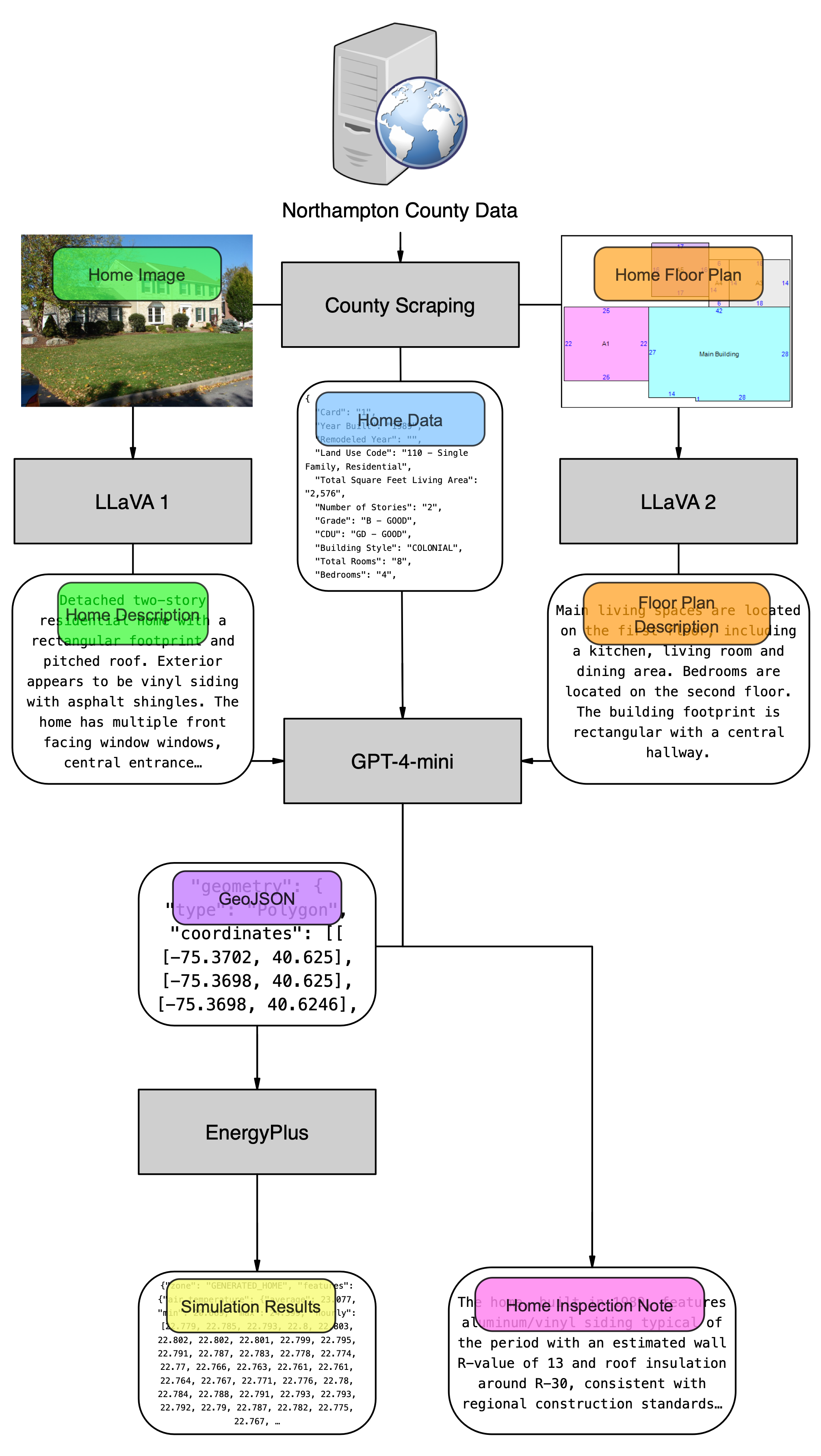}
    \caption{Overview of the pipeline.}
    \label{fig:pipeline}
\end{figure}

\rev{We introduce a reusable framework for generating simulation-ready residential datasets from publicly available information under conditions of data scarcity. The framework combines heterogeneous public records, generative \gls{ai} models, and physics-based simulation into four modular components, depicted in Figure~\ref{fig:pipeline}. Each stage focuses on a transformation of the data, from data collection to simulation:}


\begin{enumerate}
    \item A web scraper that collects data and images;
    \item An image processor powered by \gls{llava}~\parencite{liuVisualInstructionTuning2023};
    \item A \gls{geojson} and inspection note generator using \gls{gpt}~\parencite{openaiIntroducingGPT41API2025}; and
    \item An EnergyPlus\footnote{EnergyPlus\texttrademark\hspace{1mm}is a trademark of the United States Department of Energy.}~\parencite{crawleyEnergyPlusCreatingNewgeneration2001} simulation
\end{enumerate}

 While our pipeline is tailored to generate data from images and tabular data, it is inherently modular and extensible. Any stage is easily modified or expanded for other related purposes, and different information can be generated in addition to or in place of any data produced by this pipeline. For example, a different \gls{llm} could be employed in place of GPT, or other types of data could be considered or generated by the pipeline. Those interested in generating synthetic commercial buildings could use a modified version of this pipeline that emphasizes the characteristics and data central to their research. This opens use of our work to a larger set of individuals.  Rather than proposing a rigid set of steps, we present a flexible framework that integrates generative \gls{ai}, web scraping, and building simulation, designed for rapid, low-cost data generation in building energy and other fields that traditionally suffer from data scarcity. Ultimately, the data generated by our framework can be used to prototype, design, train, and research urban energy systems and models.

An instance of this pipeline was implemented and executed\footnote{Source code: \url{https://github.com/Lafayette-EshbaughSilveyra-Group/synthetic-homes}.} on a \texttt{g3.medium} flavor JetStream 2 virtual machine running Ubuntu 24.04.2 (Noble). Specifically, the system featured one NUMA node with 8 AMD EPYC-Milan processors, 30 GB of RAM, and an NVIDIA A100 SXM4 40 GB GPU (of which 10 GB was allocated). Additionally, it was fully virtualized (KVM and AMD-V). We paid \$0.73 (USD) for OpenAI requests to produce data for 247 homes---about \$0.0030 per home.\footnote{Assuming a conservative upper bound of 10 CPU minutes per home at 100 watts and electricity prices at \$0.15 per kilowatt-hour, local electricity costs are approximately \$0.0025 per synthetic home, giving a worst case total of \$0.0055 per synthetic home.} To run the pipeline, we used the PyTorch~\parencite{anselPyTorch2Faster2024} machine learning framework, and the Hugging Face Transformers library~\parencite{wolfTransformersStateoftheArtNatural2020} to load the models we used. In the subsections that follow, we describe each pipeline component in detail.

\subsection{Scraping the Data}
\label{sec:pipeline:scrape}

\begin{figure}
  \centering
  \begin{subfigure}[b]{0.45\textwidth}
    \includegraphics[width=\linewidth]{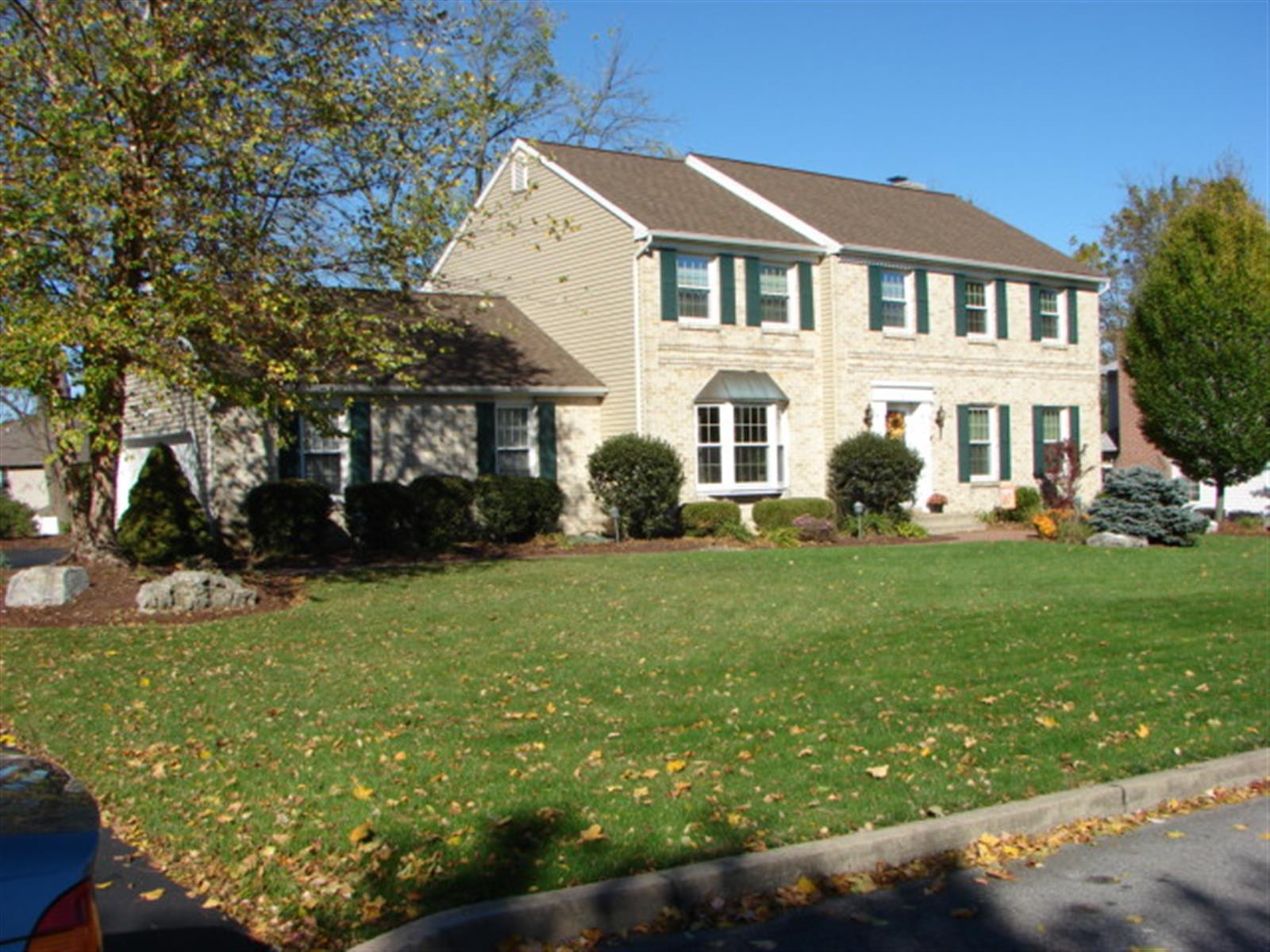}
  \end{subfigure}
  \hfill
  \begin{subfigure}[b]{0.45\textwidth}
    \includegraphics[width=\linewidth]{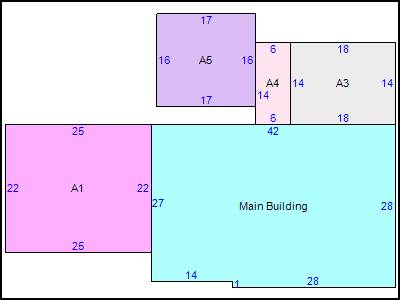}
  \end{subfigure}
  \caption{An example street view photograph and floor plan from the Northampton County database.}
  \label{fig:norco_images}
\end{figure}

To ensure the synthetic data we generated reflects real-world homes, we collected data from publicly available county datasets. For our pipeline instance, we gathered data from Northampton County, Pennsylvania, where two of the authors work and reside. To collect this data, we implemented a web scraper using \texttt{Selenium WebDriver}~\parencite{selenium} and \texttt{ChromeDriver}~\parencite{chromedriver}. For each residential property on each street in the list of streets to process, the scraper extracts key attributes, such as number of rooms, total floor area, and other metadata listed in Table~\ref{tab:metadata}. This data is saved to a \gls{json} file for each home. Additionally, the scraper downloads two images: a street view photograph and a floor plan, both of which are made available on the County's public-facing platform. An example of each is depicted in Figure~\ref{fig:norco_images}.

Although this implementation targets a single county, the pipeline is modular and can be adapted to different public record schemas and regional metadata formats. Currently, our scraper is configured to gather information from Northampton County's instance of iasWorld. Developed by Tyler Technologies, iasWorld is an Enterprise Assessment \& Tax solution used by ``jurisdictions in 26 US states, four Canadian provinces, and the Bahamas'' \parencite{tyler2022orange}. However, our implementation could easily be modified to point to another instance, such as in DeKalb County, Georgia.

Moreover, multiple web scraping modules could be employed to gather data from differently structured systems simultaneously. For example, Lancaster County, Pennsylvania uses a tool by Devnet called Edge. To create a dataset consisting of homes in Lancaster, Northampton, and DeKalb county, simply create two versions of the scrapers, one for Devnet and one for iasWorld. While much of the high-level scraping logic would remain consistent across systems, the underlying HTML structure, identifiers, and parsing rules would require tool-specific adjustments. Nevertheless, the modular design of the ingestion pipeline allows these adaptations to be incorporated without changing the overall framework.

\begin{table}
\centering
\setlength{\tabcolsep}{4pt}
\begin{tabularx}{\linewidth}{@{}lX@{}}
\toprule
\textbf{Collected Data} & \textbf{Type} \\
\midrule
Year Built & Year \\
Remodeled Year & Year \\
Land Use Code & String \\
Total Square Feet Living Area & Number \\
Number of Stories & Number \\
Grade & String \\
CDU & String \\
Building Style & String \\
Total Rooms & Number \\
Bedrooms & Number \\
Full Baths & Number \\
Half Baths & Number \\
Additional Fixtures & Number \\
Total Fixtures & Number \\
Heat/Air Cond & String \\
Heating Fuel Type & String \\
Heating System Type & String \\
Attic Code & String \\
Unfinished Area & Number \\
Rec Room Area & Number \\
Finished Basement Area & Number \\
Fireplace Openings & Number \\
Fireplace Stacks & Number \\
Prefab Fireplaces & Number \\
Basement Garage (Number of Cars) & Number \\
Condo Level & Number \\
Condo/Townhouse Type & String \\
Basement & String \\
Exterior Wall Material & String \\
Physical Condition & String \\
Sketch Data & Dictionary of floorplan segment areas \\
\bottomrule
\end{tabularx}
\caption{All data we collected from the county portal. Not all homes have values for each variable.}
\label{tab:metadata}
\end{table}
\subsection{Processing Images}
\label{sec:pipeline:proc}

After scraping the street view photographs and floor plans from the county portal, they are analyzed using \gls{llava}~\parencite{liuVisualInstructionTuning2023} version 1.5 with 7 billion parameters, distributed on HuggingFace (\texttt{llava-hf/llava-1.5-7b-hf}). This step processes both files to translate them into a textual description. For example, the floor plans can be used to define the geometry of the home, and the image to determine the quality and number of windows. 
The goal of utilizing the floor plan image is not to generate a one-to-one replication of a given home, but instead, it is used to produce structurally plausible layouts.
 This allows our pipeline to infer important aspects of the homes that will be used to generate the \gls{geojson}, which is ultimately used to generate the \gls{idf} file required by EnergyPlus.

\subsubsection{Evaluating Visual Focus with Occlusion Testing}
\label{sec:pipeline:proc:image_exp}

In image processing, it is crucial that only key portions of the images in question trigger differences in the response of a model. For example, if the prompt asks about the state of the roof of a home, the tree next to the home should be inconsequential in comparison to the roof itself. In this document, we refer to this measure as \textit{focus}. Following, we present an evaluation of focus using occlusion, which is intended to measure differential sensitivity rather than measuring semantic correctness or providing a formal visual attribution method.  

\gls{occlusion} measures the necessity of a part of an image: if removing a feature causes the prediction to change, that feature was necessary for the model's decision~\parencite{hookerBenchmarkInterpretabilityMethods2019, balkirNecessitySufficiencyExplaining2022}. When we remove image portions, we replace them with a neutral value, such as black. This creates a masked version of the image. The degree of change measured between the model's output for this masked image and its baseline response reflects the importance of the masked cell. High sensitivity to the removal of a feature suggests the model relied heavily on that region to produce its output, indicating that the region was internally necessary to the model's decision process.

During the development of our pipeline, we identified two candidate \glspl{llm} to employ for image processing. To determine the best model for this task, we compared the focus of the \gls{gpt} and \gls{llava} \glspl{llm} when analyzing images using occlusion tests. We provided the models with twenty different images of homes: ten with roofs showing no signs of damage, and ten with roofs with significant damage. Each model was instructed to evaluate and describe the state of the roof and justify its evaluation. We first took a baseline where no part of the image was masked. Then, the image was divided into one hundred cells, a number we selected to balance effectiveness, efficiency, and cost. Each cell of the image was masked, and the model was executed on the original prompt and this new image. Textual responses were embedded into vectors using the \texttt{all-MiniLM-L6-v2} SentenceTransformers model~\parencite{reimersSentenceBERTSentenceEmbeddings2019}, a lightweight and widely used model sufficient for obtaining sentence-level numerical representations. Then, to measure the semantic difference between each execution of the experiment, we took the cosine distance of each embedded response from the baseline. Additionally, heatmaps of these differences were assembled using \texttt{matplotlib}~\parencite{thematplotlibdevelopmentteamMatplotlibVisualizationPython2025}.

\begin{figure}
    \centering
    \begin{subfigure}[b]{0.45\textwidth}
        \includegraphics[width=\linewidth]{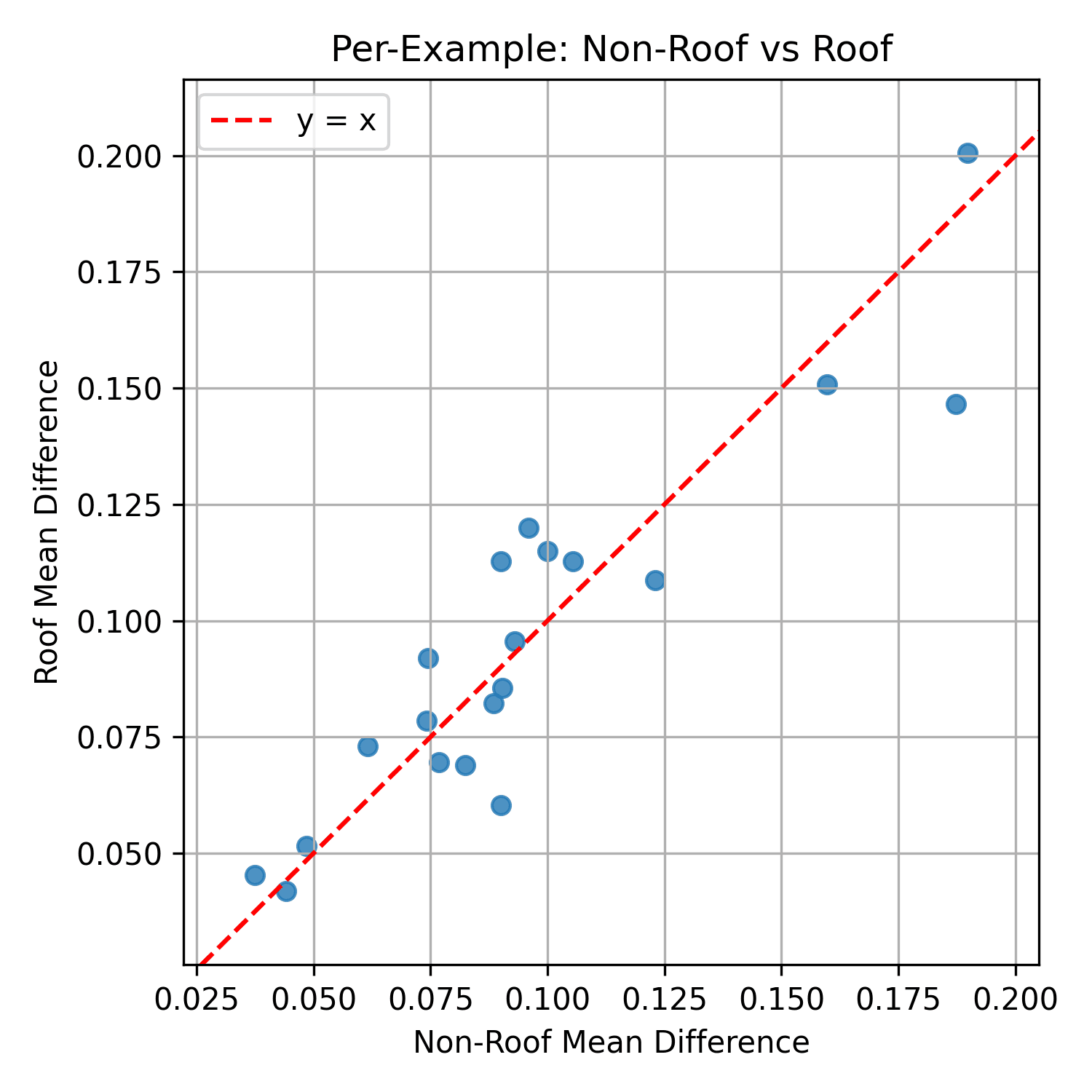}
        \caption{\gls{gpt} results.}
        \label{fig:gpt_occ_overall:scatter}
    \end{subfigure}
    \begin{subfigure}[b]{0.45\textwidth}
        \includegraphics[width=\linewidth]{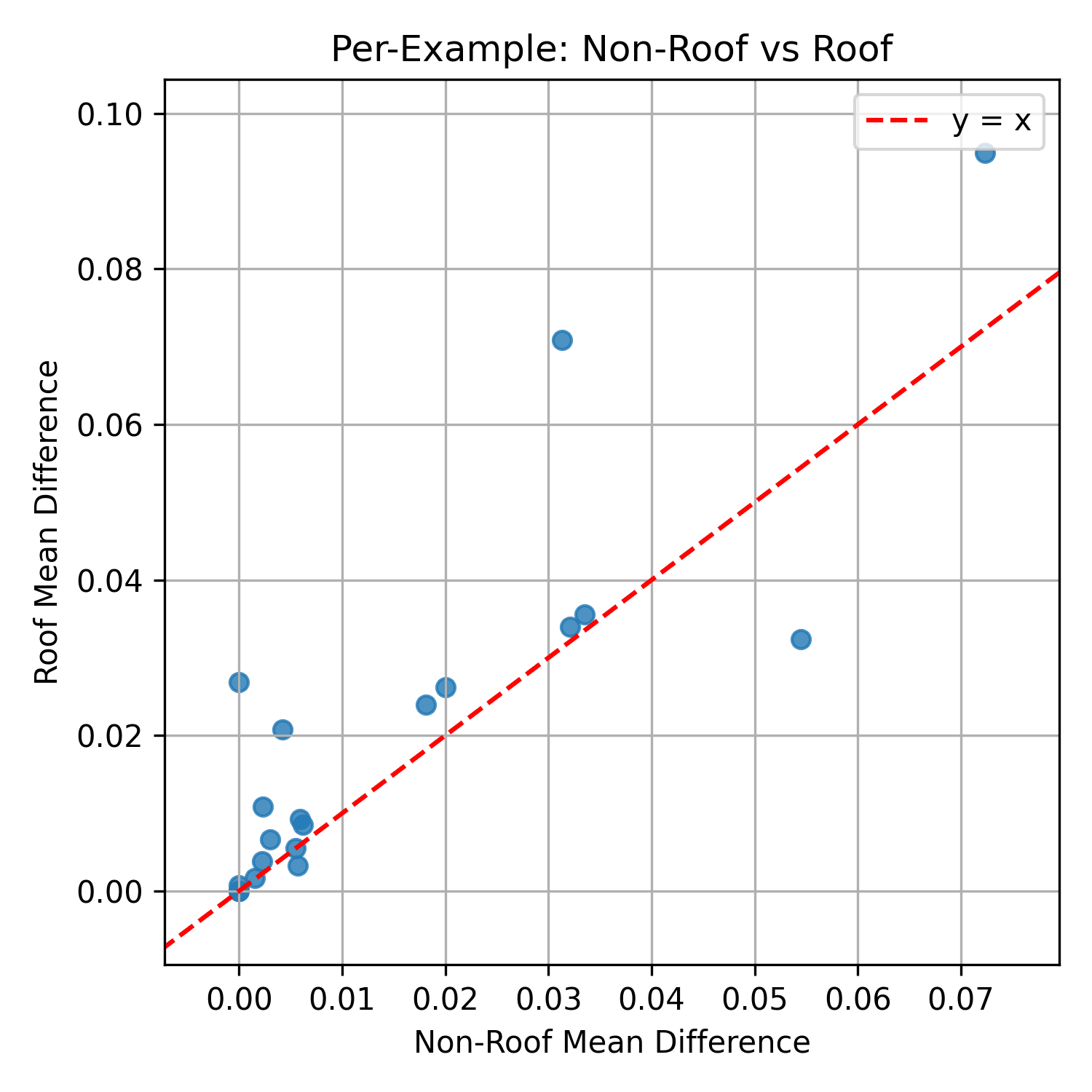}
        \caption{\gls{llava} results.}
        \label{fig:llava_occ_overall:scatter}
    \end{subfigure}
    \caption{\gls{gpt} and \gls{llava} forward occlusion per image results.}
    \label{fig:scatterplots}
\end{figure}

We categorize cells as follows: if a cell contains any visible portion of the roof, it is a roof cell; otherwise, it is considered a non-roof cell. This categorization was performed manually for all twenty images. The mean difference value in roof cells and non-roof cells for each image was calculated. Figure~\ref{fig:scatterplots} depicts the distribution of the forward occlusion results. These scatter plots are particularly telling---\gls{gpt} (Figure~\ref{fig:gpt_occ_overall:scatter}) evenly scatters the data points on either side of the line \( y = x\), indicating it has no preference for which area in the image is considered important. In comparison, \gls{llava} (Figure~\ref{fig:llava_occ_overall:scatter}) tends to produce results in the upper half of the graph. \rev{These results illustrate that, unlike \gls{gpt}, \gls{llava} consistently exhibits greater sensitivity to roof regions than to non-roof regions, indicating stronger task-relevant visual focus in our occlusion experiments.}

Additionally, overall means and standard deviations were calculated for \gls{gpt} and \gls{llava} (Figure~\ref{fig:occ_numerical_results}). In a well-focused model, the mean difference in the relevant area should be greater than the mean difference anywhere else. This behavior occurs since relevant cells should account for the majority of the response given by the model. In our experiments, we observe that \gls{gpt} reports identical roof and non-roof mean differences (Figure~\ref{fig:occ_numerical_results:gpt}). However, \gls{llava} achieved a roof mean difference approximately \( 20\% \) greater than its non-roof mean difference (Figure~\ref{fig:occ_numerical_results:llava}).

\begin{figure}
\begingroup
\setlength{\tabcolsep}{4pt}
\centering

\begin{subfigure}[b]{0.45\linewidth}
  \centering\small
  \begin{tabular}{@{}lrr@{}}
    \toprule
     & Mean Difference & SD \\
    \midrule
    Roof     & 0.0956 & 0.0384 \\
    Non-Roof & 0.0956 & 0.0408 \\
    \bottomrule
  \end{tabular}
  \caption{\gls{gpt} occlusion overall statistics}
  \label{fig:occ_numerical_results:gpt}
\end{subfigure}
\hfill
\begin{subfigure}[b]{0.45\linewidth}
  \centering\small
  \begin{tabular}{@{}lrr@{}}
    \toprule
     & Mean Difference & SD \\
    \midrule
    Roof     & 0.0208 & 0.0241 \\
    Non-Roof & 0.0149 & 0.0197 \\
    \bottomrule
  \end{tabular}
  \caption{\gls{llava} occlusion overall statistics}
  \label{fig:occ_numerical_results:llava}
\end{subfigure}

\caption{Occlusion statistics for \gls{gpt} and \gls{llava}.}
\label{fig:occ_numerical_results}
\endgroup
\end{figure}

Cells were partitioned into roof and non-roof groups within each image. The difference between the mean occlusion effects of the two groups was computed for each image and subsequently averaged across all images to obtain the mean paired difference (MPD). This value is a numerical representation of the relative importance of both cells that depict the roof and those that do not. Thus, the sign of the MPD indicates which cell group produced the larger occlusion effect, while its magnitude reflects the size of the difference. The results are reported in Table~\ref{tab:paired_occlusion_differences}.

To assess whether the MPD is statistically significant, we employed paired Wilcoxon signed-rank tests due to the small sample size and the non-normal, zero-heavy distribution of the differences. A statistically significant result (\(p < 0.05\)) indicates that the roof and non-roof mean differences are distinct for the given language model. 
GPT showed no statistically detectable difference between roof and non-roof regions (\( p=0.729 \)), while LLaVA demonstrated significantly greater roof-region sensitivity (\( p=0.0086 \)) with a moderate paired effect size (\( d_z=0.463 \)). \rev{Although occluding both roof and non-roof regions altered LLaVA's responses, the significantly \rev{larger} occlusion effects observed for roof regions indicate that the model relied more heavily on task-relevant visual information.} These statistical findings are summarized in Table~\ref{tab:paired_occlusion_differences}.

\begin{table}[ht]
\centering

\begin{tabular}{lccc}
\hline
\textbf{Model} & \textbf{Mean Paired Difference} & \textbf{Wilcoxon $p$-value} & \textbf{Cohen's $d_z$} \\
\hline
GPT   & $-0.000055$ & $0.729$  & $-0.003$ \\
LLaVA & $0.005880$  & $0.0086$ & $0.463$ \\
\hline
\end{tabular}

\caption{Paired roof-minus-non-roof occlusion differences by model. Mean paired difference is defined as \rev{roof mean difference - non-roof mean difference.}}
\label{tab:paired_occlusion_differences}
\end{table}

\begin{figure}
  \centering
  \begin{subfigure}[b]{0.45\textwidth}
    \includegraphics[width=\linewidth]{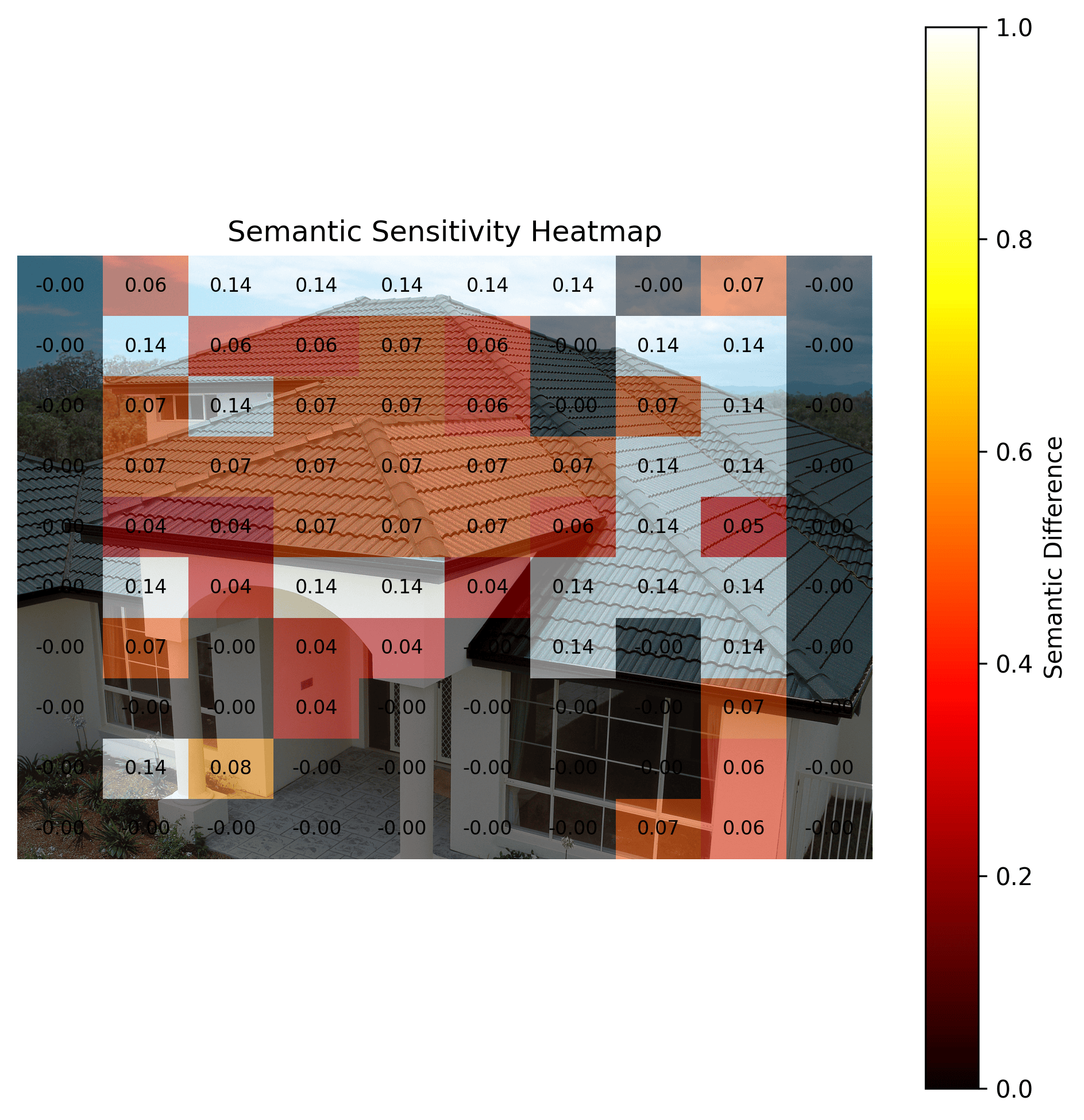}
    \caption{\gls{llava} occlusion results. These results show less sensitivity and more focus on \rev{roof cells compared to non-roof cells}.}
    \label{fig:llava_occ}
  \end{subfigure}
  \hfill
  \begin{subfigure}[b]{0.45\textwidth}
    \includegraphics[width=\linewidth]{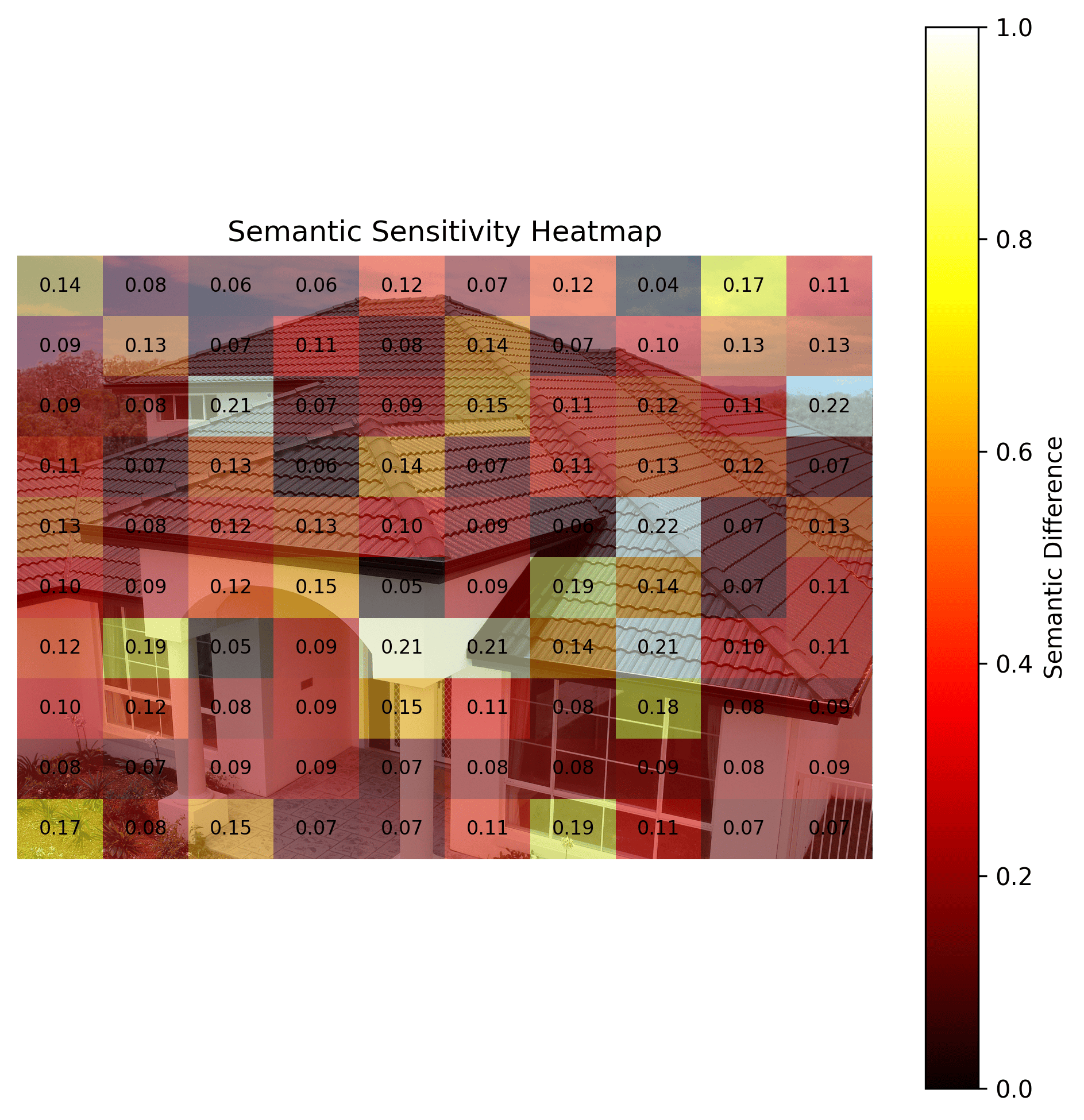}
    \caption{\gls{gpt} occlusion results. \rev{The model exhibits highly variable responses to image occlusions, indicating limited spatial selectivity between roof and non-roof regions.}}
    \label{fig:gpt_occ}
  \end{subfigure}
  \caption{\gls{gpt} and \gls{llava} occlusion results.}
  \label{fig:occ}
\end{figure}

When we compared \rev{our} individual GPT and \gls{llava} occlusion results, we found that \gls{llava} \rev{has greater roof-region sensitivity}. \gls{llava} is much more particular about which cells are key to construct the response; GPT produces results that look stochastic. Figure~\ref{fig:occ} displays one noteworthy example---\gls{llava} focuses on a more specific section of the image, primarily consisting of the home's roof. GPT, on the other hand, behaves much more randomly. Each value in the cell is the semantic difference, and the background tint of the cell represents the scaled semantic difference, where the brightest background represents the largest difference and the darkest represents the smallest. In Figure~\ref{fig:llava_occ}, the semantic difference is generally larger on the roof than on other parts of the image. A small semantic difference is a signal that a given portion of the image is less important to the model in order while formulating a response. In Figure~\ref{fig:gpt_occ}, the majority of cells are moderately important to the model in formulating a response---even cells containing walls or grass. Comparing the two, \gls{llava} demonstrates better focus on the more important parts of the image. Full occlusion experimental details, hypotheses, and findings are available in \ref{app:occlusion}.

The goal of the occlusion experiments was not to evaluate the quality of the image-to-text translation, as this aspect has been examined extensively in prior work by the model's developers~\parencite{liuVisualInstructionTuning2023, liuImprovedBaselinesVisual2024}. Those studies report strong performance on image description tasks, motivating our use of LLaVA as the image description component of the pipeline. 

\subsection{Generating GeoJSON}
\label{sec:pipeline:gen}

Once the images are processed, we utilize the descriptions given by \gls{llava} in a \gls{gpt} prompt, available in \ref{app:gpt_prompt}, to generate a \gls{geojson} file and home inspection notes. The prompt provides the image descriptions (Section~\ref{sec:pipeline:proc}) and data that was scraped from the county portal (Section~\ref{sec:pipeline:scrape}) and directs the model to take two actions: 

\begin{itemize}
\item The prompt instructs \gls{gpt} to generate a \gls{geojson} file for the building, including geometry, data from the county, and five estimated performance parameters. These estimated parameters, defined according to standard convention, include the HVAC heating and cooling coefficients of performance and \rev{R}-values for the roof and wall.

\item The prompt directs \gls{gpt} to write a short inspection note, focused on energy-related observations such as insulation, HVAC type/age, visible windows, and any inferred upgrades. This inspection note is placed in the \gls{geojson} file.
\end{itemize}

The prompt contains strict guidelines intended to improve output validity, completeness, and simulation-oriented realism. In the pipeline, generated building footprints are additionally constrained using metadata-derived target footprint area estimates based on the reported living area and number of stories for each home. The prompt explicitly instructs \gls{gpt} to generate compact residential-scale geometries whose footprint approximately matches the target estimate derived from the metadata. After generation, the \rev{area of the} resulting geometry is validated against the reported footprint area. \rev{Given that most footprints are characteristically rectangular, we validate accuracy of the generated home by comparing the area of the generated footprint with the reported values.}

If \rev{the area of} the generated footprint falls outside the acceptable scale range, defined in our experiments as between \(0.5\times\) and \(2.5\times\) the true square footage of the home, the pipeline automatically reprompts \gls{gpt} with corrective feedback regarding the size of the generated geometry. When more than six \rev{generation} attempts fail to satisfy this constraint, a deterministic scaling correction is applied to improve footprint consistency while preserving the overall building shape. \rev{Table~\ref{tab:geometry_retries} shows the number of generation attempts required for each home. Of the 247 generated homes, 116 (47.0\%) satisfied the geometric constraint on the initial generation. The remaining 131 homes required at least one regeneration, with progressively fewer requiring additional attempts. Only two homes (0.8\%) exhausted all six generation attempts and required deterministic scaling.}

\begin{table}[t]
\centering
\revtable{
\begin{tabular}{lrrr}
\toprule
\textbf{Generation Attempts} & \textbf{Homes} & \textbf{Percent} & \textbf{Cumulative Homes} \\
\midrule
1 (accepted immediately) & 117 & 47.4\% & 117 \\
2 & 70 & 28.3\% & 187 \\
3 & 35 & 14.2\% & 222 \\
4 & 14 & 5.7\% & 236 \\
5 & 6 & 2.4\% & 242 \\
6 & 3 & 1.2\% & 245 \\
Deterministic scaling & 2 & 0.8\% & 247 \\
\midrule
\textbf{Total} & \textbf{247} & \textbf{100.0\%} & \textbf{247} \\
\bottomrule
\end{tabular}
}
\caption{\rev{Number of generation attempts required before satisfying the area footprint constraint. Only two homes required deterministic scaling after exhausting all six generation attempts.}}
\label{tab:geometry_retries}
\end{table}

To evaluate scale consistency between the synthetic homes and the original residential metadata, we compared generated footprint areas against county-derived target footprint estimates computed from the reported living area and number of stories for each home. Generated footprint areas exhibited a statistically significant positive correlation with these metadata-derived target areas (\(r = 0.3383, p < 0.001; R^2 = 0.1145\)).

While the previous analysis evaluates the plausibility of the \rev{area of the} generated homes, it does not address the reliability of the underlying structured-data generation process. We use \gls{gpt} for structured data generation given its strong ability to produce well-formed \gls{json} outputs from descriptive prompts. Prior work has demonstrated that \glspl{llm} can reliably generate outputs that conform to predefined schemas under strict schema constraints, even with no previous prompting or task-specific fine tuning. For example, \rev{the} StructuredRAG benchmark suite showed \( 82\% \) format compliance under \gls{json} prompt conditions~\parencite{shortenStructuredRAGJSONResponse2024}. 

As we built the pipeline, we reviewed \gls{gpt}'s responses and found that it seldom failed to adhere to the expected \gls{json} structure. In the rare case that malformed output is produced, the pipeline catches the parsing error and automatically reprompts \gls{gpt}. Only one synthetic home failed generation, \( 0.403 \% \) of the attempted homes. An excerpt from an example \gls{geojson} object is provided in Figure~\ref{fig:json_example}.

\begin{figure}
    \centering
    \begin{minipage}{0.9\linewidth}
    \begin{lstlisting}[language=json]
{
  "type": "Feature",
  "properties": {
    "name": "Generated Home",
    "floor_area": 2576,
    "building_type": "Single family",
    "inspection_note": "...",
    "hvac_heating_cop": 0.85,
    "hvac_cooling_cop": 3.0,
    "wall_r_value": 13,
    "roof_r_value": 30,
  },
  "geometry": {
    "type": "Polygon",
    "coordinates": [...]
  }
}
    \end{lstlisting}
    \end{minipage}
    \caption{Excerpt of a generated \gls{geojson} file.}
    \label{fig:json_example}
\end{figure}

\subsection{Running EnergyPlus Simulations}
\label{sec:pipeline:eplus}

To execute a simulation in EnergyPlus, we take the \gls{geojson} generated by \gls{gpt} (Section~\ref{sec:pipeline:gen}) and convert it to an \gls{idf} file for use in EnergyPlus. We use a template \gls{idf} file, filled with default values that are replaced with variables from the \gls{geojson} data, such as the geometry, HVAC heating and cooling coefficients of performance and \rev{R}-values for the wall and roof. Additionally, the template has many fields with constant values, such as the simulation running period and materials, constructions, and zones used in the simulation. Once the details from the \gls{geojson} are copied into the template, a final \gls{idf} file is produced. 

The templating process is facilitated with the \texttt{eppy} package~\parencite{santoshphilipEppy2024}, which makes programmatic \gls{idf} generation straightforward. Then, we use ExpandObjects to prepare the file for simulation. ExpandObjects is bundled with EnergyPlus, and its primary use case is to take an \gls{idf} and expand template objects---such as the \texttt{HVACTemplate} we use---into full systems~\parencite{expandobjects2023}. Finally, we run the expanded \gls{idf} file through EnergyPlus, producing final simulation results.

We write the inspection notes and EnergyPlus simulation results for each example into one \gls{jsonl} file. This completes our end-to-end dataset generation process, yielding a unified dataset that combines textual and simulation-based information. 

\section{Validating \rev{Plausibility} of Pipeline Generated Data}
\label{sec:validating_data}

\begin{revision}
Following prior work on synthetic data evaluation~\parencite{patkiSyntheticDataVault2016, goncalvesGenerationEvaluationSynthetic2020, duSystematicAssessmentTabular2025, lautrup2024syntheval}, we view synthetic data plausibility as application dependent, linked to the properties of the domain in focus rather than as a single universal mechanism to validate data across all domains. Given that the purpose of our framework is to generate simulation-ready residential building models and in this field there is not an agreed validation approach for such synthetic data, we utilize interval coverage metrics to evaluate the empirical plausibility of our generated data.

Interval coverage metrics are widely used to evaluate whether model outputs remain within statistically expected ranges~\parencite{krugerPredictionIntervalsEconomic2024,Schmidinger2023}. \textcite{krugerPredictionIntervalsEconomic2024} use interval coverage as a principal measure of interval calibration, whereas \textcite{Schmidinger2023} evaluate coverage alongside complementary uncertainty metrics. Furthermore, percentile distributions have been used as part of the evaluation of synthetic building-energy profiles~\parencite{Patidar2016}. Drawing on these complementary evaluation perspectives, we define plausible reference intervals of the corresponding ResStock~\parencite{ResStock2024_1Dataset, ResStockTechRefV3_3_0} variables and evaluate the proportion of generated values that fall within these intervals. ResStock is a publicly available dataset developed by the U.S. Department of Energy to represent the U.S. residential building stock. The selected interval, the 10th--90th percentiles, excludes the extreme tails of the empirical ResStock distribution while retaining its central region. The bottom of the interval was selected given that \textcite{Patidar2016} leave the 0th-10th percentiles out of their evaluation interval, and the top was selected for symmetry.
\end{revision}

\rev{Our comparison is restricted to Climate Zone 4A and to the four variables that vary across synthetic homes---Wall R-Value, Roof R-Value, Cooling COP, and Heating COP---because of their importance in building energy simulation. Each variable is evaluated separately against its corresponding ResStock reference interval. This reflects the objective of assessing the plausibility of individual generated building characteristics, including uncommon but physically plausible homes that need not conform to the joint distribution of the reference dataset.} The remaining approximately 150 variables are fixed template values outside the generative pipeline, \rev{although} the pipeline could readily be extended to generate these as well. The evaluation corpus consists of 247 synthetic homes and 2,000 ResStock Climate Zone 4A samples. In addition to \rev{the proposed plausibility coverage metric}, we report \rev{statistics that provide descriptive context} for both datasets.

Table~\ref{tab:realism_results} reports the \rev{plausibility coverage} results for the four variables across all synthetic homes by displaying the percentage of values that fall within the \rev{ResStock 10th--90th percentile interval}. We observed coverage greater than \(58\%\) for all variables and greater than \(95\%\) for three of the four variables. Under our \rev{coverage}-based definition of \rev{plausibility}, these results suggest that the generated values for these variables are \rev{largely consistent with the
typical ranges observed in the reference dataset.} Additionally, the table reports summary statistics for both datasets, including the mean and median of each variable for the synthetic homes and the interquartile range (IQR) observed in ResStock. These statistics are provided for \rev{descriptive} context and are not \rev{used when computing the plausibility coverage metric.}

\begin{table}[t]
    \centering
    \small
    \begin{tabular}{c|cccc}
    \toprule
    Variable & \rev{Coverage} (\%) & ResStock Q1--Q3 & Synthetic Mean & Synthetic Median \\
    \midrule
    Wall R-Value
        & 98.38
        & 1.94--2.64
        & 2.42
        & 2.29 \\
    Roof R-Value
        & 100.0
        & 3.35--6.69
        & 5.21
        & 5.28 \\
    Cooling COP
        & 58.70
        & 3.1--3.8
        & 2.90
        & 3.00 \\
    Heating COP
        & 99.59
        & 0.8--1.0
        & 0.82
        & 0.85 \\
    \bottomrule
    \end{tabular}
    \caption{
    Coverage of synthetic homes within the \rev{ResStock 10th--90th percentile interval for each variable, together with the} ResStock interquartile range (Q1--Q3) and the synthetic homes mean and median.
    }
    \label{tab:realism_results}
\end{table}


Compared to prior generative urban-energy workflows, our framework incorporates both multimodal sensitivity analysis and \rev{reference-based plausibility} validation against a national reference dataset. Occlusion provides empirical evidence that the image processing component of the pipeline behaves as intended. \rev{We additionally show that the generated variables generally fall within the empirically plausible reference intervals derived from ResStock, with coverage exceeding 95\% for three of the four evaluated variables.} \rev{While concerns regarding hallucinations and reliability remain important when using \glspl{llm}, the results presented here indicate that the proposed safeguards and validation procedures} help mitigate some of the issues identified by \textcite{liuLargeLanguageModels2025}.

\section{Conclusion}
\label{sec:conclusion}

In this work, we presented a multimodal framework for generating simulation-ready synthetic residential data from public records, imagery \rev{by applying} generative AI models. The results demonstrate the feasibility of generating simulation-ready synthetic residential data when comprehensive real-world data are unavailable. To the best of our knowledge, no prior approach has combined these heterogeneous data sources into a unified pipeline capable of producing both structured simulation inputs and complementary textual descriptions for residential buildings. 

The proposed pipeline produces both structured building-energy models and accompanying textual descriptions, enabling the creation of dual-modality datasets suitable for simulation and downstream machine learning applications. To evaluate the reliability of the generated outputs, we introduced a validation framework that combines multimodal sensitivity analysis, structured-output verification, and plausibility-oriented comparison against a national reference dataset. Together, these analyses provide evidence that the generated homes are physically plausible, simulation compatible, and aligned with the intended objectives of the pipeline.

Recent reviews have highlighted the growing relevance of generative AI in building energy research while also identifying challenges related to reliability, validation, and practical deployment, motivating the development of robust methods for energy simulation workflows~\parencite{xuLeveragingGenerativeAI2024, liuLargeLanguageModels2025}. The framework presented in this work contributes to this emerging area by providing a flexible approach for generating simulation-ready residential data under conditions of limited data availability. Beyond its use for synthetic home generation, the framework may support a variety of downstream applications and can be extended in several directions. One particularly promising avenue \rev{for future work} is the use of the generated data to support the training of machine learning models for urban energy applications. For instance, the pipeline could be expanded to utilize the resulting simulation data and inspection notes to label the efficiency of particular components that affect the energy efficiency of a home. These labeled homes could be used to train an \gls{ml} model that provide recommendations for energy efficiency retrofits. Furthermore, such a model could help prioritize retrofit interventions for residential buildings. \rev{Moreover, such downstream use would complement the validation presented in this work by displaying the dataset's usefulness for predictive modeling.}

Overall, this work evaluates plausibility, simulation compatibility, and task-aligned multimodal sensitivity, which align with the intended use of the generated homes as simulation-ready synthetic residential data. Rather than reconstructing ground-truth building geometries or calibrated utility measurements, the pipeline is designed to produce physically plausible and operationally useful representations under conditions of limited real-world data availability. Constructing large-scale authoritative datasets that satisfy realism constraints remains a broader challenge in residential energy modeling and synthetic urban data generation.

Ultimately, the high cost, limited availability, and privacy concerns associated with collecting urban energy data motivate the use of synthetic alternatives. Our framework helps address these data-access challenges by producing simulation-ready synthetic data accompanied by plausibility-oriented validation. By reducing dependence on costly or restricted data sources, the approach will broaden access to data-driven research in the urban energy domain. 

\section*{Acknowledgments}
The authors extend gratitude to Lafayette College for supporting this work by funding it through the EXCEL Scholars program and providing computational resources with the Firebird high performance computing cluster. Access to the JetStream2 system was made possible via the NSF-funded ACCESS program, facilitated by Lafayette College. We also thank the NSF and ACCESS for supporting this resource, and Indiana University for hosting JetStream2 as a part of the ACCESS program.

\section*{CRediT Author Attributions
}
\textbf{Jackson Eshbaugh}: Software, Investigation, Writing - Original Draft, Visualization, Validation, Data Curation. \textbf{Chetan Tiwari}: Conceptualization, Writing - Review and Editing, Supervision (supporting). \textbf{Jorge Silveyra}: Conceptualization, Writing - Review and Editing, Methodology, Supervision (lead), Project Administration, Validation

\printglossaries
\printbibliography

\appendix
\section{Occlusion Results}
\label{app:occlusion}

In this appendix, we provide detailed information and results from our occlusion tests. Firstly, when running the test we utilized identical prompts for each model. For GPT, we prompted:

\begin{quote}
    You are a certified home inspector. Describe the status roof. Is it in good condition? Why or why not?
\end{quote}

Likewise, we prompted LLaVA with the following:

\begin{quote}
    USER: <image>\\You are a certified home inspector. Describe the status roof. Is it in good condition? Why or why not? ASSISTANT:
\end{quote}

These prompts differ slightly due to how the models are trained; namely, LLaVA expects an image token and labeled portions of the message. The majority of LLMs behave this way, and the GPT API likely wraps the given prompt in something similar.

Next, we provide our hypotheses pertaining to the performance of both models. We hypothesize that \gls{llava}'s advantage stems from its architecture: it employs an early fusion design in which images are converted into visual tokens by CLIP~\parencite{radfordLearningTransferableVisual2021} that are then prepended to the textual prompt~\parencite{liuVisualInstructionTuning2023}. These visual tokens are passed through the transformer layers of the \gls{llm} together with text tokens, enabling multimodal attention at every layer. It has been demonstrated that vision tokens are most critical for a model in early layers, when they fuse into text tokens~\parencite{zhangLLaVAMiniEfficientImage2025}. A likely result of this is the reduction of the model's sensitivity to variations in the image, which may explain its stronger performance in our experiments.

While GPT-4V (\gls{gpt}'s vision model) is proprietary, the case can be made that it possesses a different structure. Evaluation studies show that it performs well across visual reasoning and captioning tasks, but it falls short in fine grained visual problems, requiring complex visual tasks to be broken up into interleaved subfigures and text~\parencite{yangDawnLMMsPreliminary2023}. These evaluations are similar to evaluations from late fusion (connector) models. For example, BLIP-2 shows strengths in captioning images and instruction-following, while it is less robust with fine-grained spatial reasoning~\parencite{liBLIP2BootstrappingLanguageImage2023}. Another model, MiniGPT-4, demonstrates a similar profile: it has strengths in tasks such as image description, but falters with complex spatial reasoning~\parencite{zhuMiniGPT4EnhancingVisionLanguage2023}. Both BLIP-2 and MiniGPT-4 are late fusion models, where compressed visual features are routed into the \gls{llm} at later stages~\parencite{liBLIP2BootstrappingLanguageImage2023, zhuMiniGPT4EnhancingVisionLanguage2023}. As shown, BLIP-2, MiniGPT-4, and GPT-4V all suffer from a very similar weakness: fine-grained or complex spatial reasoning. Based on this similarity, we hypothesize that GPT-4V is also a late fusion model. This helps explain why \gls{gpt} performed differently from \gls{llava} in our experiments: as structure dictates function, differences in behavior strongly suggest differences in design. On the basis of this analysis, we feel confident to select \gls{llava} to complete this task.

In Table~\ref{tab:forward_occ_results}, we provide our forward occlusion results, including roof mean difference (RMD) and non-roof mean difference (NRMD) values from both GPT and LLaVA and references to figures displaying GPT and LLaVA heatmaps side by side.

\begin{landscape}
\begin{table}
\centering
\setlength{\tabcolsep}{4pt}
\small
\begin{tabularx}{\linewidth}{@{}l*{10}{c}@{}}
\toprule
\textbf{Metric} & \textbf{1} & \textbf{2} & \textbf{3} & \textbf{4} & \textbf{5} & \textbf{6} & \textbf{7} & \textbf{8} & \textbf{9} & \textbf{10} \\
\midrule
\multicolumn{11}{@{}l}{\textit{Damaged Roofs}} \\
GPT RMD           & 0.0856 & 0.0920 & 0.0730 & 0.0689 & 0.1127 & 0.0453 & 0.0516 & 0.0603 & 0.1149 & 0.0419 \\
GPT NRMD          & 0.0903 & 0.0745 & 0.0615 & 0.0824 & 0.0900 & 0.0375 & 0.0485 & 0.0900 & 0.1000 & 0.0442 \\
LLaVA RMD         & 0.0033 & 0.0324 & 0.0949 & 0.0093 & 0.0262 & 0.0269 & 0.0356 & 0.0055 & 0.0038 & 0.0208 \\
LLaVA NRMD        & 0.0057 & 0.0545 & 0.0723 & 0.0059 & 0.0200 & 0.0000 & 0.0335 & 0.0055 & 0.0022 & 0.0042 \\
Roof Cells        & 70     & 80     & 53     & 83     & 97     & 96     & 80     & 96     & 91     & 88     \\
Non-Roof Cells    & 30     & 20     & 47     & 17     & 3      & 4      & 20     & 4      & 9      & 12     \\
Figure       & \ref{fig:occ_forward_bad_1} & \ref{fig:occ_forward_bad_2} & \ref{fig:occ_forward_bad_3} & \ref{fig:occ_forward_bad_4} & \ref{fig:occ_forward_bad_5} & \ref{fig:occ_forward_bad_6} & \ref{fig:occ_forward_bad_7} & \ref{fig:occ_forward_bad_8} & \ref{fig:occ_forward_bad_9} & \ref{fig:occ_forward_bad_10} \\
\midrule
\multicolumn{11}{@{}l}{\textit{Undamaged Roofs}} \\
GPT RMD           & 0.1128 & 0.0785 & 0.1465 & 0.1508 & 0.0822 & 0.0956 & 0.1087 & 0.2005 & 0.0696 & 0.1200 \\
GPT NRMD          & 0.1055 & 0.0742 & 0.1872 & 0.1598 & 0.0884 & 0.0930 & 0.1230 & 0.1898 & 0.0768 & 0.0959 \\
LLaVA RMD         & 0.0709 & 0.0085 & 0.0007 & 0.0000 & 0.0340 & 0.0108 & 0.0066 & 0.0000 & 0.0017 & 0.0240 \\
LLaVA NRMD        & 0.0313 & 0.0062 & 0.0000 & 0.0000 & 0.0321 & 0.0023 & 0.0030 & 0.0000 & 0.0015 & 0.0181 \\
Roof Cells        & 53     & 47     & 68     & 49     & 81     & 39     & 70     & 60     & 47     & 73     \\
Non-Roof Cells    & 47     & 53     & 32     & 51     & 19     & 61     & 30     & 40     & 53     & 27     \\
Figure       & \ref{fig:occ_forward_good_1} & \ref{fig:occ_forward_good_2} & \ref{fig:occ_forward_good_3} & \ref{fig:occ_forward_good_4} & \ref{fig:occ_forward_good_5} & \ref{fig:occ_forward_good_6} & \ref{fig:occ_forward_good_7} & \ref{fig:occ_forward_good_8} & \ref{fig:occ_forward_good_9} & \ref{fig:occ_forward_good_10} \\
\bottomrule
\end{tabularx}
\caption{Quantitative results from our forward occlusion experiments, including roof mean difference (RMD) and non-roof mean difference (NRMD) for both GPT and LLaVA. The number of cells in the image considered roof and non-roof are listed, and figures of side by side comparisons of the resulting heatmaps are referenced.}
\label{tab:forward_occ_results}
\end{table}
\end{landscape}

\foreach \n in {1,...,10} {%
  \begin{figure}[t]
    \centering
    \begin{subfigure}[b]{0.48\textwidth}
      \centering
      \includegraphics[width=\linewidth]{bad_roof_\n_heatmap_gpt.png}
      \caption{GPT Heatmap}
    \end{subfigure}\hfill
    \begin{subfigure}[b]{0.48\textwidth}
      \centering
      \includegraphics[width=\linewidth]{bad_roof_\n_occlusion_heatmap_llava.png}
      \caption{LLaVA Heatmap}
    \end{subfigure}
    \caption{Forward Occlusion---Damaged Roof \n}
    \label{fig:occ_forward_bad_\n}
  \end{figure}
}

\foreach \n in {1,...,10} {%
  \begin{figure}[t]
    \centering
    \begin{subfigure}[b]{0.48\textwidth}
      \centering
      \includegraphics[width=\linewidth]{good_roof_\n_heatmap_gpt.png}
      \caption{GPT Heatmap}
    \end{subfigure}\hfill
    \begin{subfigure}[b]{0.48\textwidth}
      \centering
      \includegraphics[width=\linewidth]{good_roof_\n_occlusion_heatmap_llava.png}
      \caption{LLaVA Heatmap}
    \end{subfigure}

    \caption{Forward Occlusion---Undamaged Roof \n}
    \label{fig:occ_forward_good_\n}
  \end{figure}
}

\clearpage
\pagebreak

\section{GeoJSON Generation Prompt}
\label{app:gpt_prompt}

In this appendix, we provide the prompts used to generate the GeoJSON data
in Section~\ref{sec:pipeline:gen}. First, we list the main prompt, then the retry prompt for unreasonable geometry.

\lstset{
  basicstyle=\ttfamily\small,
  breaklines=true,
  columns=fullflexible,
  frame=single
}

\begin{lstlisting}
    You are a certified home energy inspection expert and data specialist building synthetic training data for an AI model.
    
    You are provided with:
    - Structured residential property data (JSON).
    - A detailed exterior description of the home: "{exterior_description}"
    - A detailed floorplan description: "{floorplan_description}"
    
    Your tasks:
    1. Generate a **GeoJSON file** for this building with:
    - A plausible (longitude, latitude) location in Bethlehem, PA.
    - A "FeatureCollection" containing exactly **one Feature**.
    - Geometry: Polygon or MultiPolygon representing an approximate simulation-ready building footprint.
    - The geometry should be realistic in scale for the reported square footage, number of stories, and building style.
    - Use the floorplan description to infer approximate shape, but use the structured property data to control scale.
    - Target approximate footprint area: {target_footprint_text}.
    - The generated polygon should represent the building footprint, not the parcel or lot boundary.
    - The polygon area should be within approximately 20-30% of the target footprint area when possible.
    - Avoid large coordinate spans. For typical residential homes in Bethlehem, PA, longitude/latitude differences should usually be very small.
    - Properties from the provided JSON:
        - "Year Built"
        - "Total Square Feet Living Area"
        - "Building Style"
        - "Exterior Wall Material"
        - "Heating Fuel Type"
        - "Heating System Type"
        - "Heat/Air Cond"
        - "Bedrooms"
        - "Full Baths"
        - "Half Baths"
        - "Basement"
        - "Number of Stories"
        - "Grade"
    - Estimated performance parameters:
        - "hvac_heating_cop" (0-1)
        - "hvac_cooling_cop"
        - "wall_r_value"
        - "roof_r_value"
        - "air_change_rate" (0-1)
    
    2. Write a short **inspection note** as if you had toured the home, focusing on energy-related observations: insulation, HVAC type/age, visible windows, and any inferred upgrades.
    
    **Strict Guidelines:**
    - Only base your outputs on the provided data and descriptions.
    - Do not invent details not clearly supported by the inputs.
    - Ensure the GeoJSON is valid and realistic.
    - Coordinates should place the home plausibly in Bethlehem, PA.
    - Do not generate parcel-sized rectangles.
    - Do not use coordinate differences such as 0.0005 degrees unless the resulting footprint area is consistent with the target footprint area.
    - Prefer compact residential-scale polygons.
    
    Here is the structured property data:
    
    {json.dumps(house_data)}
    
    **Output Format:**
    Return a raw JSON object:
    {{
      "geojson": {{
        "type": "FeatureCollection",
        "features": [
          {{
            "type": "Feature",
            "geometry": {{ ... }},
            "properties": {{
              ...,
              "air_change_rate": ...,
              "hvac_heating_cop": ...,
              "hvac_cooling_cop": ...,
              "wall_r_value": ...,
              "roof_r_value": ...,
            }}
          }}
        ]
      }},
      "inspection_note": "..."
    }}
    No backticks or explanation.
\end{lstlisting}

\begin{lstlisting}
    {original_prompt}
    IMPORTANT CORRECTION:
    Your previous GeoJSON geometry was rejected because its footprint area was not realistic in scale.

    Target footprint area: {last_error.get('target_area_ft2', 'unknown')} ft2
    Generated footprint area: {last_error.get('generated_area_ft2', 'unknown')} ft2
    Generated/target area ratio: {last_error.get('ratio', 'unknown')}

    If the generated/target area ratio is greater than 1, the footprint is too large; reduce the coordinate span.
    If the generated/target area ratio is less than 1, the footprint is too small; increase the coordinate span.
    Adjust the footprint scale while preserving a compact residential building shape.

    Regenerate the full raw JSON object, but correct the GeoJSON geometry so that:
    - the footprint area is approximately consistent with the target footprint area,
    - the geometry represents the building footprint, not the parcel or lot boundary,
    - the coordinates remain plausible for Bethlehem, PA,
    - the output remains valid raw JSON with the same top-level structure.
\end{lstlisting}

\end{document}